\documentclass[10pt,twocolumn,letterpaper]{article}

\usepackage{iccv}              

\definecolor{iccvblue}{rgb}{0.21,0.49,0.74}
\usepackage[pagebackref,breaklinks,colorlinks,allcolors=iccvblue]{hyperref}
\usepackage{multirow}
\usepackage{amssymb}
\usepackage{mathtools}
\usepackage{array}
\usepackage{tcolorbox}
\usepackage{enumitem}
\usepackage{listings}
\usepackage{algorithm}
\usepackage{algpseudocode}
\usepackage{xcolor}
\usepackage[utf8]{inputenc}
\usepackage{amsmath}
\usepackage{hyperref}

\def\paperID{1141} 
\def\confName{ICCV}
\def\confYear{2025}

\title{ReAL-AD: Towards Human-Like Reasoning in End-to-End Autonomous Driving}

\author{
    \textbf{
        Yuhang Lu$^{1}$, 
        Jiadong Tu$^{1}$, 
        Yuexin Ma$^{1,\dagger}$,
        Xinge Zhu$^{2,\dagger}$ 
    } \\
    $^{1}$ ShanghaiTech University \\
    $^{2}$ The Chinese University of Hong Kong \\
    {\tt\small \{luyh2, tujd2023, mayuexin\}@shanghaitech.edu.cn, zhuxinge123@gmail.com}
}

\begin{document}
\maketitle

\footnotetext[\value{footnote}]{\textdagger{} Corresponding authors.}

\begin{abstract}
   End-to-end autonomous driving has emerged as a promising approach to unify perception, prediction, and planning within a single framework, reducing information loss and improving adaptability. However, existing methods often rely on fixed and sparse trajectory supervision, limiting their ability to capture the hierarchical reasoning process that human drivers naturally employ. To bridge this gap, we propose ReAL-AD, a Reasoning-Augmented Learning framework that structures decision-making in autonomous driving based on the three-tier human cognitive model: \textbf{Driving Strategy}, \textbf{Driving Decision}, and \textbf{Driving Operation}, where Vision-Language Models (VLMs) are incorporated to enhance situational awareness and structured reasoning across these levels. Specifically, we introduce: (1) the \textbf{Strategic Reasoning Injector}, which formulates high-level driving strategies by interpreting complex traffic contexts from VLM-generated insights; (2) the \textbf{Tactical Reasoning Integrator}, which refines strategic intent into interpretable tactical choices such as lane changes, overtaking, and speed adjustments; and (3) the \textbf{Hierarchical Trajectory Decoder}, which progressively translates tactical decisions into precise control actions for smooth and human-like trajectory execution. Extensive evaluations show that integrating our framework improves planning accuracy and safety by over 30\%, making end-to-end autonomous driving more interpretable and aligned with human-like hierarchical reasoning. 
The project page can be found at: \href{https://4dvlab.github.io/project_page/realad}{\texttt{4dvlab.github.io/project\_page/realad}}

\end{abstract}

\section{Introduction}
\label{sec:intro}
End-to-end autonomous driving~\cite{hu2022st,jiang2023vad,weng2024drive, hu2023planning, zheng2024genad, chen2024vadv2, sun2024sparsedrive} has gained significant attention due to its advantages, such as minimizing information loss and simplifying system architecture. However, existing methods face fundamental limitations. Most notably, they rely on fixed and sparse trajectory supervision, which fails to replicate the structured cognitive reasoning inherent to human drivers – a process involving hierarchical understanding of contextual relationships (e.g., traffic rules, interactive intentions) and adaptive decision-making hierarchy.

\begin{figure}
    \centering
    \includegraphics[width=\linewidth]{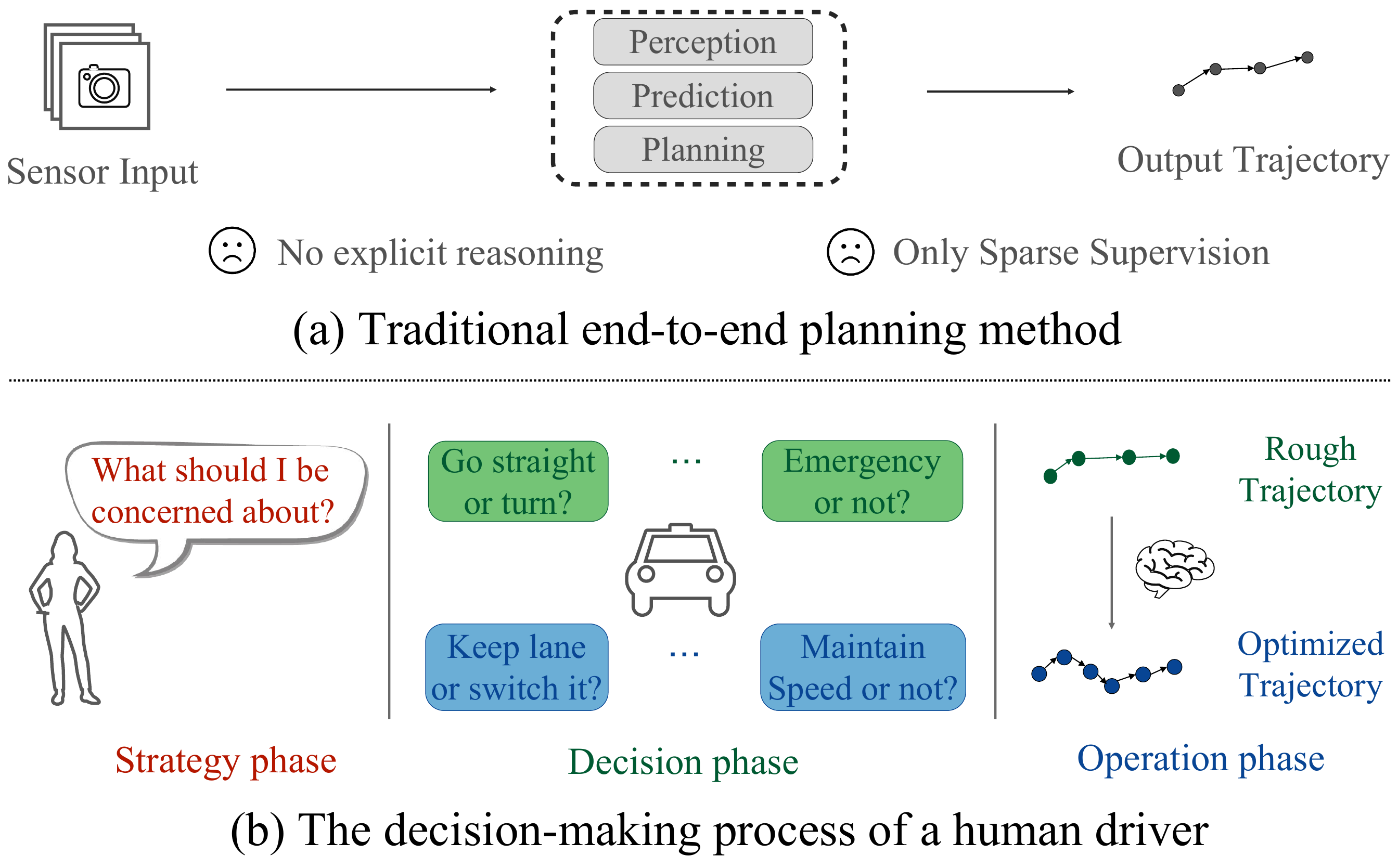}
    \caption{\textbf{Comparison between end-to-end networks and human driving logic.} (a) illustrates the workflow and limitations of the end-to-end autonomous driving system, while (b) depicts the structured decision-making process of a human driver. }  
    \label{fig:teaser}
    \vspace{-4mm}
\end{figure}

Recent efforts~\cite{xu2024drivegpt4, mao2023language, wang2023drivemlm, tian2024drivevlm, pan2024vlp, xu2024vlm} to incorporate Vision-Language Models (VLMs) \cite{achiam2023gpt, deepseek-llm, bai2023qwen, wang2023cogvlm, liu2023llava} aim to address these challenges by leveraging their semantic reasoning capabilities and cross-modal alignment. Most studies treat VLMs as auxiliary modules, providing additional semantic cues to enhance perception and planning. These approaches treat semantic reasoning as \textbf{an isolated preprocessing step} rather than embedding it organically into the \textbf{decision hierarchy}, where strategic planning, tactical decision-making, and operational control work in synergy. This lack of structured multi-stage reasoning hinders generalization and limits the effectiveness of current VLM-driven solutions in diverse real-world scenarios.

So how do Human Drivers perform the decision-making? As illustrated in Fig.~\ref{fig:teaser}, human drivers make decisions in a structured, hierarchical manner, progressively refining their actions from high-level reasoning to precise execution.
At the \textbf{Driving Strategy level}, drivers analyze the overall driving scene to establish a strategic understanding. This involves identifying key traffic elements, prioritizing relevant objects, and incorporating contextual factors such as traffic rules and road conditions. This strategic awareness provides the foundation for subsequent decision-making.
At the \textbf{Driving Decision level}, drivers translate their strategic understanding into tactical choices, such as determining whether to maintain their lane, overtake, or adjust speed. These decisions are made dynamically based on traffic flow, surrounding vehicles, and predicted interactions, ensuring safe and efficient maneuvering.
At the \textbf{Driving Operation level}, drivers refine tactical decisions into precise vehicle control actions. This includes adjusting steering, throttle, and braking to execute the intended maneuvers smoothly, conducting lateral control actions including lane changes to maintain stability and safety.
However, most existing end-to-end autonomous driving systems fail to explicitly model this hierarchical cognitive process. Instead, they often rely on direct trajectory prediction without structured reasoning, leading to suboptimal planning and a lack of human-like decision-making granularity.

Building upon this insight, we propose \textbf{ReAL-AD}, a novel \textbf{Re}asoning-\textbf{A}ugmented \textbf{L}earning framework for \textbf{A}utonomous \textbf{D}riving, which leverages the reasoning capabilities of Vision-Language Models (VLMs) to embed human-like hierarchical decision-making into end-to-end autonomous driving systems. Inspired by the structured cognitive process of human drivers, ReAL-AD introduces a multi-level architecture that explicitly models decision-making from strategic reasoning to tactical execution, thereby enhancing interpretability and adaptability in dynamic environments. 

Specifically, the \textbf{Strategic Reasoning Injector} extracts high-level situational awareness from VLM-generated insights, encoding them into ego-centric queries to guide downstream planning. The \textbf{Tactical Reasoning Integrator} refines these high-level decisions into structured, interpretable tactical control commands, ensuring both feasibility and consistency with real-world driving constraints. Finally, the \textbf{Hierarchical Trajectory Decoder} employs a hierarchical variational decoder to iteratively refine trajectory planning in a two-stage process, ensuring coherence between high-level strategic intent and low-level tactical execution. By integrating VLM-driven reasoning across all levels of decision-making, ReAL-AD significantly improves interpretability, adaptability, and overall driving performance in complex and dynamic traffic scenarios.

Our extensive experiments on the NuScenes~\cite{caesar2020nuscenes} and Bench2Drive~\cite{jia2024bench2drive} datasets demonstrate the superior performance of our approach over existing methods. Compared to the baseline, our method achieves a 33\% reduction in L2 error and a 32\% decrease in collision rate, significantly improving trajectory accuracy and driving safety across diverse dynamic scenarios. Additionally, a comprehensive ablation study systematically evaluates each component's contribution, further validating our framework's effectiveness.  Our contributions can be summarized as follows:
\begin{itemize}
\item[$\bullet$] We propose \textit{ReAL-AD}, a novel reasoning-augmented end-to-end autonomous driving framework that explicitly incorporates hierarchical decision-making and aligns trajectory planning with human cognitive processes. 
\item[$\bullet$] We introduce the Strategic Reasoning Injector for VLM-driven decision integration, the Tactical Reasoning Integrator for structured control, and the Hierarchical Trajectory Decoder for hierarchical trajectory refinement, ensuring consistency from reasoning to execution.
\item[$\bullet$] Our framework, when integrated into baseline approaches, demonstrates over 30\% improvement in both trajectory planning accuracy and safety metrics across NuScenes and Bench2Drive datasets, while ablation studies confirm the contribution of each component.
\end{itemize}

\section{Related Work}
\label{sec:related}
\subsection{End-to-End Autonomous Driving}
End-to-end autonomous driving planning methods \cite{hu2022st, hu2023planning, jiang2023vad, chen2024vadv2, zheng2024genad, weng2024drive, liao2024diffusiondrive, chen11ppad, sun2024sparsedrive, li2024enhancing, guo2024end, doll2024dualad, li2024does} achieve their ultimate goal by simultaneously training multiple modules, reducing information loss during the pipeline process, and making them a popular area of research. ST-P3 \cite{hu2022st} introduces a design that integrates multiple auxiliary information or tasks to enhance planning performance. UniAD \cite{hu2023planning} demonstrates impressive performance in open-loop evaluations. Subsequently, VAD \cite{jiang2023vad} introduces compact vectorized scene representations, which not only improve planning efficiency but also reduce computational costs. PARA-Drive \cite{weng2024drive} investigates the impact of the design order of auxiliary tasks within an end-to-end framework. GenAD \cite{zheng2024genad} models autonomous driving as a future generation problem, conducting motion prediction and ego planning simultaneously within a structured latent trajectory space. Following this, works such as VADv2 \cite{chen2024vadv2} and DiffusionDrive \cite{liao2024diffusiondrive} explore the integration of probabilistic modeling into planning to enhance accuracy.

However, these methods rely on a fixed set of future ego-vehicle trajectories as ground truth, leading to sparse supervision. This is problematic since human drivers use complex decision-making, which ego-vehicle trajectories alone fail to capture. This limits an autonomous driving system’s learning and generalization. In contrast, our approach leverages Vision-Language Models (VLMs) for a deeper understanding of driving behavior. By integrating high-dimensional strategies and tactical commands while predicting trajectories with finer granularity through a hierarchical decoder, the network mimics human reasoning, enhancing planning decisions and overall autonomous driving performance.

\begin{figure*}[t]
  \includegraphics[width=\linewidth]{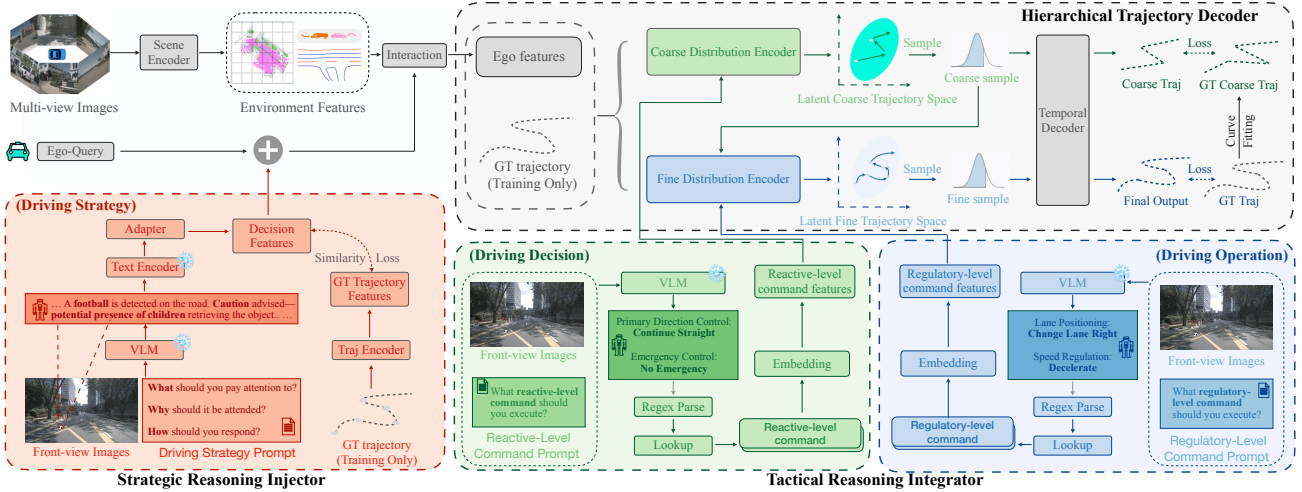}
  \caption{\textbf{Overall pipeline of ReAL-AD. }Multi-view images are processed by the Scene Encoder to extract environmental features. The Strategic Reasoning Injector generates driving decisions using structured prompts and utilizes them to enhance ego-query. The Tactical Reasoning Integrator outputs reactive- and regulatory-level command features, which are then fed into the Hierarchical Trajectory Decoder. This decoder progressively refines the latent trajectory space to generate the final planning trajectory.}
\label{fig:pipeline}
\vspace{-4mm}
\end{figure*}

\subsection{VLMs for Autonomous Driving}
Recently, Vision-Language Models (VLMs) have demonstrated remarkable performance across a wide range of general tasks \cite{achiam2023gpt, team2023gemini, deepseek-llm, bai2023qwen, wang2023cogvlm, liu2023llava, yu2024rlaifv, li2023blip2bootstrappinglanguageimagepretraining}, and their integration into autonomous driving systems~\cite{chen2023sharegpt4v, li2023monkey, young2024yi, glm2024chatglm, tian2024tokenize, ma2024dolphins, lu2025can} has become a prominent area of research. A number of studies \cite{chen2024driving, xu2024drivegpt4, sha2023languagempc, mao2023language, wang2023drivemlm, huang2024drivegpt, wen2023dilu, hwang2024emma, xing2024openemma, mao2023gpt, liu2023mtd, wang2024omnidrive, sima2024drivelm, shao2024lmdrive, huang2024making, qian2024fasionad} treat VLMs as agents that receive driving scene images and text prompts as inputs, generating driving decisions as outputs. For instance, Drive-with-LLMs \cite{chen2024driving} encode the perception information into a latent space, which is then fed into a large language model (LLM) to predict future planning trajectories. DriveGPT4 \cite{xu2024drivegpt4} takes front-camera video inputs, employing VLMs to predict control signals for planning and provide decision explanations. LanguageMPC \cite{sha2023languagempc} converts historical ground truth perception data and HD maps into language format, using chain-of-thought reasoning to analyze driving scenes and generate planning actions. AgentDriver \cite{mao2023language} converts driving situations into textual descriptions with human-like intelligence, then uses an LLM to reason and plan. Additionally, DriveMLM \cite{wang2023drivemlm} validates the effectiveness of VLM-based planning models within a closed-loop simulation environment. However, while VLMs can capture complex visual and language cues, they often lack a full grasp of the underlying physics and constraints of driving behavior and limited 3D spatial understanding, resulting in less accurate or unsafe trajectory predictions compared to end-to-end systems that consider more comprehensive environmental feedback.

An alternative approach leverages the decision-making capabilities of VLMs as extra inputs to end-to-end autonomous driving systems. This approach primarily focuses on utilizing VLM outputs to refine or guide the learning process of these systems. DriveVLM \cite{tian2024drivevlm} employs VLMs as a slower system to generate driving trajectories, with complementary networks acting as references when needed. VLM-AD \cite{xu2024vlm} uses the VLM as a teacher to generate free-form reasoning and construct action annotations, thereby assisting in the learning of the end-to-end network. VLP \cite{pan2024vlp} enhances autonomous driving systems by strengthening their understanding of vehicle contexts and environments. Senna \cite{jiang2024senna} decouples high-level planning from low-level trajectory prediction, creating a more modular and interpretable planning framework. Existing methods incorporate decision features via distillation and contrastive learning but treat semantic reasoning as a separate preprocessing step, limiting generalization in real-world scenarios. To address this, we leverage VLM to generate driving strategy and tactical commands and introduce a hierarchical trajectory decoder to translate these decisions into precise control actions, simulating the human driving thought process.

\section{Methods}
\label{sec:methods}
\subsection{Overview}
In this section, we present a VLM-assisted framework for human-like reasoning-augmented learning, integrating a three-tier human cognitive model—\textbf{Driving Strategy}, \textbf{Driving Decision}, and \textbf{Driving Operation}—into end-to-end autonomous driving systems. First, we revisit the operation paradigm of conventional end-to-end planning systems (Sec.~\ref{subsec: preliminary}). To inject human-like reasoning patterns, we develop: (1) Strategic Reasoning Injector (Sec.~\ref{subsec: high-level-decision}) formulates high-level driving strategies by interpreting complex traffic contexts from VLM-generated insights; (2) Tactical Reasoning Integrator (Sec.~\ref{subsec: low-level-command}) refines strategic intent into interpretable tactical choices; (3) Hierarchical Trajectory Decoder (Sec.~\ref{subsec: hierarchical-decoder}) mimics human intuition-refinement patterns via hierarchical planning — first establishing rough motion patterns, then refining detailed trajectories. Finally, in Sec.~\ref{subsec: loss-function}, we introduce how to supervise the network.

\subsection{Preliminary}
\label{subsec: preliminary}
In a vision-based end-to-end autonomous driving system, multi-view camera image streams are first processed by an image backbone (e.g., ResNet \cite{he2016deep}) to extract 2D visual features. These features are then transformed into 3D or Bird’s Eye View (BEV) scene representations via view transformation modules, capturing scene geometry such as road layouts, agent positions, and lane topology. Transformer-based task-specific decoders enable task queries (e.g., agent queries, map queries) to interact with scene features, modeling relevant representations for planning. The planning module initializes an ego-query embedding, which attends to scene features and task queries through cross-attention layers to construct ego-features, enabling the ego vehicle to contextualize its state. Finally, the trajectory decoder, typically a Multi-Layer Perceptron (MLP), predicts future waypoints by regressing from the ego-features.

\begin{figure}[t]
  \includegraphics[width=\linewidth]{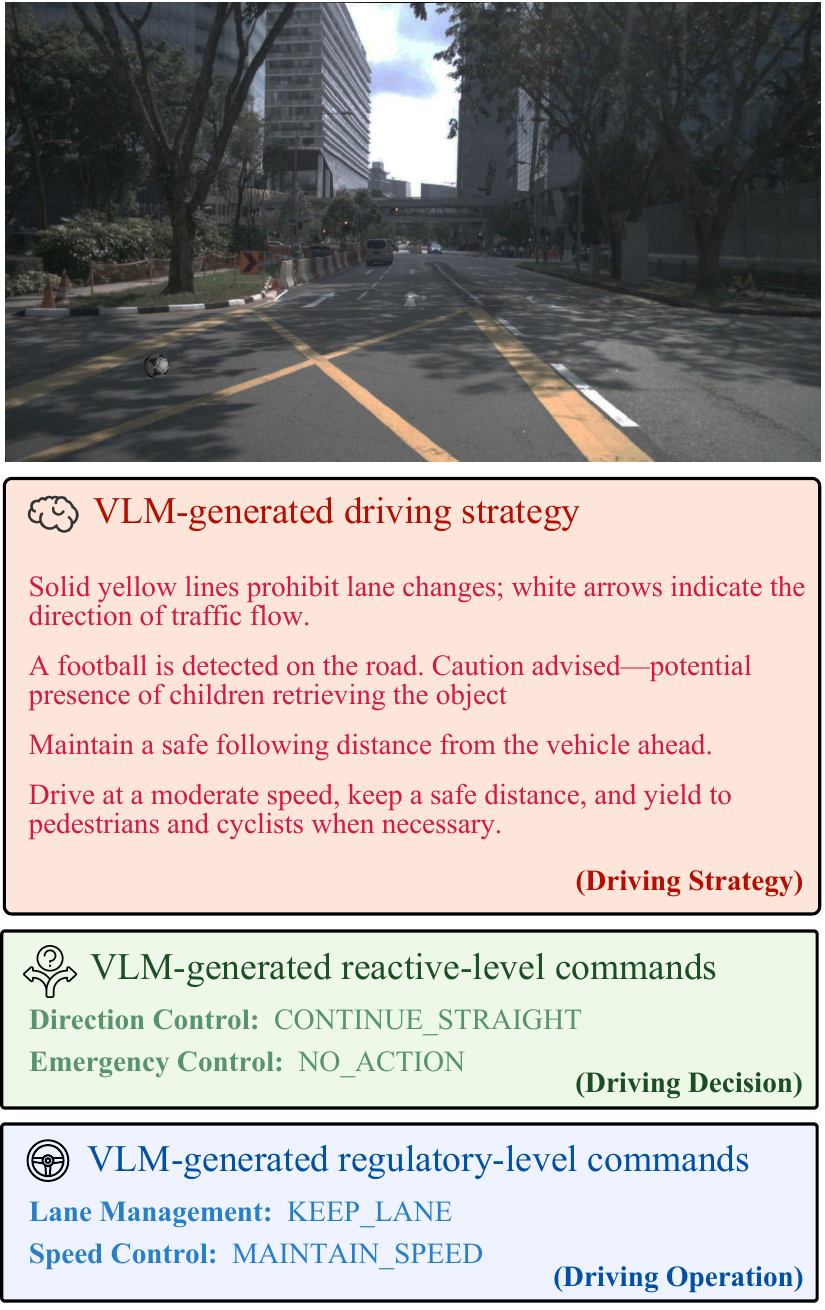}
  \vspace{-5mm}
  \caption{\textbf{Example VLM output}. Given a front-view image and prompt, VLM generates corresponding driving strategy, reactive-level and regulatory-level commands. }
\label{fig:vlm_outputs}
\vspace{-9mm}
\end{figure}

\subsection{Strategic Reasoning Injector}
\label{subsec: high-level-decision}

When navigating a driving scenario, human drivers first make \textbf{driving strategies} by identifying critical traffic participants or relevant regulations—cognitive processes that we replicate using VLM-generated reasoning. These structured insights are then encoded to guide the ego-query, which serves as the starting point of the planning module.

The VLM first generates textual driving strategy through prompt-guided visual reasoning:
\begin{equation} 
T_{strategy} = \underbrace{M(V, P_{strategy})}_{\text{VLM-based driving strategy generation}}
\end{equation}
where $M(\cdot)$ denotes the VLM processor, $V$ the visual input, and $P_{strategy}$ denotes the driving strategy prompt template.

The generated strategy text $T_{strategy}$ is then encoded into semantic space through a pretrained language encoder $E_{text}(\cdot)$. To bridge the modality gap between linguistic strategy and visual perceptions, we choose a lightweight adapter module implemented as:
\begin{equation} 
F_{strategy} = MLP(E_{text}(T_{strategy}))
\end{equation}

Concurrently, we encode the ground-truth trajectory $gt_{traj}$ into planning features $F_{gt}$ using trajectory encoder $E_{gt}(\cdot)$. A cosine similarity loss enforces consistency between strategy semantics and planning dynamics:
\begin{equation} 
\mathcal{L}_{sim} = 1 - \cos(F_{strategy}, F_{gt})
\end{equation}

The refined strategy features are then integrated into the ego-query through residual updating:
\begin{equation} 
Q_{ego} \leftarrow Q_{ego} + F_{strategy}
\end{equation}

This approach injects VLM-derived reasoning capabilities into ego-query, thereby guiding the acquisition of key planning-related features when interacting with environmental features. More details can be found in Appendix

\subsection{Tactical Reasoning Integrator}
\label{subsec: low-level-command}
While driving strategies provide semantic context, their abstract nature (\textit{e.g.},``yield to approaching vehicles") may lack actionable controls for trajectory planning. To bridge this gap, we introduce \textbf{tactical command} that translates semantic strategies into executable choices, establishing dual-level reasoning from strategic planning to tactical driving decision and operation.

The VLM generates structured commands through categorical-constrained visual reasoning:
\begin{equation}
T_{command} = \underbrace{M(V, P_{command})}_{\text{VLM-based structured command generation}}
\end{equation}
where $M(\cdot)$ denotes the VLM processor, $V$ the visual input, and $P_{command}$ denotes the command prompt template and enforces four-category output format (Direction/Emergency/Lane/Speed).

The raw text output $T_{command}$ undergoes deterministic parsing to extract executable commands:
\begin{equation}
\{c_k\}_{k=1}^4 = \text{RegexParse}\left( T_{command}, 
    \begin{bmatrix}
    \mathcal{P}_{dir} \\ 
    \mathcal{P}_{lane} \\
    \mathcal{P}_{speed} \\
    \mathcal{P}_{emerg}
    \end{bmatrix} \right)
\end{equation}
where $\mathcal{P}_*$ are pre-defined regular expression patterns for each category. Each text command $c_k$ is then mapped to its category-specific index:
\vspace{-1mm}
\begin{equation}
i_k = \text{Lookup}(\mathcal{D}_k, c_k), \quad \mathbf{c}_k = \text{onehot}(i_k) \in \{0,1\}^{n_k}
\end{equation}

with $\mathcal{D}_k$ being the command dictionary for category $k$ containing all predefined options. This two-step conversion guarantees machine-readability while preserving VLM's semantic understanding.

The discrete commands undergo category-specific encoding and strategic fusion:
\vspace{-1mm}
\begin{equation}
\begin{cases}
\mathbf{e}_{dir} = E_{dir}(\mathbf{c}_1), \quad \mathbf{e}_{emerg} = E_{emerg}(\mathbf{c}_4) \\
\mathbf{e}_{lane} = E_{lane}(\mathbf{c}_2), \quad \mathbf{e}_{speed} = E_{speed}(\mathbf{c}_3)
\end{cases}
\end{equation}

where $\{E_k\}$ are learnable embedding matrices. The encoded features are then partitioned by vehicle control level:

\begin{equation}
\begin{aligned}
& \mathbf{g}_{reactive} = \mathbf{e}_{dir} \oplus \mathbf{e}_{emerg} \quad  \\
& \mathbf{g}_{regulatory} = \mathbf{e}_{lane} \oplus \mathbf{e}_{speed} \quad 
\end{aligned}
\end{equation}

$\mathbf{g}_{reactive}$  encodes the driving decision at the instantaneous reaction level, while $\mathbf{g}_{regulatory}$ encodes the driving operation command after careful consideration. These two commands function at different layers within the subsequent hierarchical trajectory decoder, offering detailed and precise guidance to facilitate the generation of human-like trajectories from coarse to fine.

\subsection{Hierarchical Trajectory Decoder}
\label{subsec: hierarchical-decoder}
Inspired by the hierarchical nature of human driving cognition—where \textbf{instantaneous reactions occur before deliberate adjustments}, we propose a two-layer variational decoder with latent trajectory spaces conditioned on control at both the driving decision and the driving operation level.

The first-layer decoder models rough motion patterns using ego-vehicle features and reactive-level tactical commands (directional intent and emergency indicators) via conditional variational inference:

\begin{enumerate}
    \item \textbf{Distribution Parameterization}: Map inputs to latent coarse trajectory space. 
    \begin{equation}
        \mu^c, \log\sigma^c = \text{DistributionEncoder}_\theta\big([f_{ego}; \mathbf{g}_{reactive}])
    \end{equation}
    
    \item \textbf{Latent Sampling}: Extract global motion patterns.
    \begin{equation}
        z^c \sim \mathcal{N}(\mu^c, \sigma^c) \quad 
    \end{equation}
\end{enumerate}

where $f_{ego} \in \mathbb{R}^d$ is ego-vehicle features, $\mathbf{g}_{reactive} \in \mathbb{R}^{16}$ denotes reactive-level command features, and coarse latent code $z^c \in \mathbb{R}^k$ captures rough motion pattern.

The second layer refines the coarse motion pattern from the first layer by incorporating multi-source conditioning and mapping it to a fine-grained trajectory representation.
\begin{enumerate}
    \item \textbf{Hierarchical Refinement}: Condition latent fine trajectory space on coarse motion patterns and regulatory-level command.
    \begin{equation}
        \mu^f, \log\sigma^f = \text{DistributionEncoder}_\phi\big([f_{ego};z^c;\mathbf{g}_{regulatory}]\big)
    \end{equation}
    
    \item \textbf{Latent Sampling}: Sample fine-grained trajectory representation.
    \begin{equation}
        z^f \sim \mathcal{N}(\mu^f, \sigma^f) \quad 
    \end{equation}
\end{enumerate}

where $\mathbf{g}_{regulatory} \in \mathbb{R}^{16}$ denotes regulatory-level command features which encode lane management and speed control choices.

After obtaining latent representations $z^c$ and $z^f$ from the hierarchical variational process, the next step is to decode these latent variables into actual trajectory sequences. Our trajectory decoding extends the temporal modeling in \cite{zheng2024genad} with dual latent-stream processing. The process is:
\vspace{-1mm}
\begin{equation}
\begin{aligned}
&\text{\textbf{Coarse Stream}:} \quad & h^{c} &= \text{GRU}(\text{repeat}(z^c, T), f_{ego}) \\
& & pred_{coarse} &= \text{MLP}(h^c) \\
&\text{\textbf{Fine Stream}:} \quad & h^{f} &= \text{GRU}(\text{repeat}(z^f, T), f_{ego}) \\
& & pred_{fine} &= \text{MLP}(h^f) \\
\end{aligned}
\end{equation}

where $z^c$ and $z^f$ represent the approximate motion pattern and the refined trajectory representation, respectively. $T$ denotes the number of future timesteps for planning. $h^c$ and $h^f$ correspond to the latent feature sequences for coarse and fine trajectory prediction. $pred_{coarse}$ and $pred_{fine}$ denote the planned coarse and fine trajectories, respectively.

\subsection{Loss Function}
\label{subsec: loss-function}
Our composite loss function integrates five key components for effective hierarchical learning:

\begin{equation}
\begin{aligned}
\mathcal{L}_{\text{overall}} = &\underbrace{\mathcal{L}_{baseline}}_{\text{Base model}} + 
\underbrace{\lambda_0\mathcal{L}_{sim} + \lambda_1\mathcal{L}_{gt\_recon}}_{\text{Strategic Reasoning Injection Loss}} \\
&+ \underbrace{\lambda_2\mathcal{L}_{coarse} + \lambda_3\mathcal{L}_{hier\_kl}}_{\text{Hierarchical Trajectory Decoder Loss}}
\end{aligned}
\end{equation}

\paragraph{Base Model. } 
We retain all baseline model losses, denoted as $\mathcal{L}_{\text{baseline}}$.

\paragraph{Strategic Reasoning Injection Loss. } 
For the Strategic Reasoning Injection Module, we introduce two losses:
\begin{enumerate}
    \item \textbf{$\mathcal{L}_{sim}$} enforces consistency between strategy semantics and planning dynamics, as defined in~\ref{subsec: high-level-decision}
    
    \item \textbf{$\mathcal{L}_{gt\_recon}$} supervises GT trajectory encoding to ensure proper feature representation:
    \begin{equation}
        \mathcal{L}_{gt\_recon} = \| \text{Dec}(F_{gt}) - \text{gt}_{traj} \|_2
    \end{equation}
    where $\text{Dec}(\cdot)$ denotes our hierarchical trajectory decoder.
\end{enumerate}

\vspace{-2ex}

\paragraph{Hierarchical Trajectory Decoder Loss. } 
To ensure effective hierarchical trajectory planning, we introduce two key loss functions for the Hierarchical Trajectory Decoder:

\begin{enumerate}
    \item $\mathcal{L}_{\text{coarse}}$ applies the baseline objective to the predicted coarse trajectory, $pred_{coarse}$, and the Bézier curve-fitted \cite{pastva1998bezier} coarse trajectory ground truth $gt_{coarse}$. 
    \begin{equation}
        \mathcal{L}_{coarse} = \mathcal{L}_{baseline}(pred_{coarse}, gt_{coarse})
    \end{equation}

    \item $\mathcal{L}_{\text{hier-kl}}$ enforces consistency across the hierarchical latent spaces using a two-level KL divergence formulation:
    \begin{equation}
    \mathcal{L}_{\text{KL}} = \beta_c \cdot \underbrace{D_{KL}(q_{\phi}(z^c_{\text{curr}} \| z^c_{\text{fut}}))}_{\text{Coarse-level}}
    + \beta_f \cdot \underbrace{D_{KL}(q_{\phi}(z^f_{\text{curr}} \| z^f_{\text{fut}}))}_{\text{Fine-level}}
    \end{equation}
    
    For hierarchy level $l \in \{c, f\}$, the KL divergence is computed as:
    \vspace{-2mm}
    \begin{equation}
        D_{KL}^l = \frac{1}{2} \left( \log \frac{\sigma^2_{\text{fut},l}}{\sigma^2_{\text{curr},l}} - 1 
        + \frac{\sigma^2_{\text{curr},l} + (\mu_{\text{curr},l} - \mu_{\text{fut},l})^2}{\sigma^2_{\text{fut},l}} \right)
    \end{equation}
    
    The overall hierarchical KL loss is then computed as:
    \begin{equation}
        \mathcal{L}_{\text{hier}} = D_{KL}^c + D_{KL}^f
    \end{equation}
\end{enumerate}

\section{Experiments}
\label{sec:exp}
\begin{table*}[ht]
\centering
\begin{tabular}{l|cccc|cccc}
\toprule
\multirow{2}{*}{Model}          & \multicolumn{4}{c|}{L2(m) $\downarrow$} & \multicolumn{4}{c}{Collision Rate (\%) $\downarrow$} \\
                                & 1s    & 2s   & 3s   & Avg. & 1s     & 2s     & 3s     & Avg.   \\ \midrule
NMP \cite{zeng2019end}                            & 0.53  & 1.25 & 2.67 & 1.48 & 0.04   & 0.12   & 0.87   & 0.34   \\
FF \cite{hu2021safe}                             & 0.55  & 1.20 & 2.54 & 1.43 & 0.06   & 0.17   & 1.07   & 0.43   \\
EO \cite{khurana2022differentiable}                             & 0.67  & 1.36 & 2.78 & 1.60 & 0.04   & 0.09   & 0.88   & 0.33   \\ \midrule
ST-P3 \cite{hu2022st}                         & 1.33  & 2.11 & 2.90 & 2.11 & 0.23   & 0.62   & 1.27   & 0.71   \\
UniAD$\dagger$ \cite{hu2023planning}                          & 0.48  & 0.96 & 1.65 & 1.03 & 0.05   & 0.17   & 0.71   & 0.31   \\
OccNet$\dagger$ \cite{tong2023scene}                         & 1.29  & 2.13 & 2.99 & 2.14 & 0.21   & 0.59   & 1.37   & 0.72   \\
VAD-Base \cite{jiang2023vad}                       & 0.41  & 0.70 & 1.05 & 0.72 & 0.07   & 0.17   & 0.41   & 0.22   \\
PARA-Drive$\dagger$  \cite{weng2024drive}                    & 0.40  & 0.77 & 1.31 & 0.83 & 0.07   & 0.25   & 0.60   & 0.30   \\
GenAD   \cite{zheng2024genad}                        & 0.28  & 0.49 & 0.78 & 0.52 & 0.08   & 0.14   & 0.34   & 0.19   \\
UAD \cite{guo2024end}                            & 0.39  & 0.81 & 1.50 & 0.90 & \textbf{0.01}   & 0.12   & 0.43   & 0.19   \\ \midrule
DriveVLM*  \cite{tian2024drivevlm}                     & 0.15  & 0.29 & 0.48 & 0.31 & 0.05   & 0.08   & 0.17   & 0.10   \\
VLP-UniAD  \cite{pan2024vlp}                     & 0.36  & 0.68 & 1.19 & 0.74 & 0.03   & 0.12   & 0.32   & \underline{0.16}   \\
VLP-VAD \cite{pan2024vlp}                        & \underline{0.30}  & 0.53 & 0.84 & 0.55 & \textbf{0.01}   & \textbf{0.07}   & 0.38   & \textbf{0.15}\\
VLM-AD (UniAD) \cite{xu2024vlm}                 & 0.39  & 0.82 & 1.43 & 0.88 & 0.05   & 0.11   & 0.43   & 0.19   \\
VLM-AD (VAD) \cite{xu2024vlm}                   & \textbf{0.24}  & \textbf{0.46} & 0.75 & \textbf{0.48} & 0.12   & 0.17   & 0.41   & 0.23   \\
Senna  \cite{jiang2024senna}                         & 0.37  & 0.54 & 0.86 & 0.59 & 0.09   & 0.12   & 0.33   & 0.18   \\ \midrule
Ours (VAD \& MiniCPM-V)  & \underline{0.30}  & \underline{0.48} & \textbf{0.67} & \textbf{0.48} & 0.07   & \underline{0.10}   &\textbf{ 0.28}& \textbf{0.15}   \\
Ours (UniAD \&. MiniCPM-V)$\dagger$      &  0.40     & 0.71     & 1.14     & 0.77     & \underline{0.02}       & 0.12       &  0.37      &  0.17      \\
Ours (VAD \& Qwen-VL) & 0.35      &  0.53    & \underline{0.71}     & \underline{0.53}     &  0.09      & 0.12       & \underline{0.31}       & 0.17       \\
Ours (UniAD \& Qwen-VL)$\dagger$ &  0.39     & 0.75     & 1.09     & 0.74     & \underline{0.02}        &    0.14    &  0.39      &   0.17     \\ \bottomrule
\end{tabular}
\vspace{-2mm}
\caption{\textbf{Open-loop planning results on nuScenes dataset.} ``*'' indicates ego-vehicle status features are used as input. ``$\dagger$ denotes results evaluated using UniAD metrics. ''The best result is in \textbf{bold}, while the second-best is \underline{underlined}. (Methods that use ego-vehicle status features as input are not considered in this ranking)}
\label{tab:nuscenes_comparison}
\vspace{-3mm}
\end{table*}

\begin{table*}[ht]
\centering
\resizebox{\linewidth}{!}{
\begin{tabular}{l|cccccccc|cc}
\toprule
\multirow{3}{*}{Model} &
  \multicolumn{8}{c|}{Open-loop Metric} &
  \multicolumn{2}{c}{Closed-loop Metric} \\ \cmidrule{2-11} 
 &
  \multicolumn{4}{c|}{L2(m) $\downarrow$} &
  \multicolumn{4}{c|}{Collision Rate (\%) $\downarrow$} &
  \multirow{2}{*}{Driving Score $\uparrow$} &
  \multirow{2}{*}{Success Rate(\%) $\uparrow$} \\
                           & 1s   & 2s   & 3s   & \multicolumn{1}{c|}{Avg.} & 1s   & 2s   & 3s   & Avg. &                    &                   \\ \midrule
UniAD-tiny \cite{jiang2023vad}                & 0.51 & 0.91 & 1.37 & \multicolumn{1}{c|}{0.93} & 0.84 & 2.29 & 3.44 & 2.19 & 32.00              & 9.54              \\
UniAD-base  \cite{jiang2023vad}               & 0.38 & 0.72 & 1.12 & \multicolumn{1}{c|}{0.74} & 0.54 & 1.65 & 3.36 & 1.85 & 37.72              & 9.54              \\
VAD  \cite{jiang2023vad}                      & 0.45 & 0.91 & 1,47 & \multicolumn{1}{c|}{0.94} & 0.10 & 0.20 & 0.29 & 0.20 & 39.42              & 10.00             \\
AD-MLP \cite{zhai2023rethinking} & 1.98     &  3.63    &  5.29    & \multicolumn{1}{c|}{3.39} & 5.92     & 8.03     &  9.70    &  7.88    & 17.91                   &   0.00                \\
TCP   \cite{wu2022trajectory}      & 1.25     & 1.70     & -     & \multicolumn{1}{c|}{1.48}     &  3.31    & 3.00     & -     & 3.15     &     32.43               &   9.54                \\ \midrule
Ours (VAD \& MiniCPM-V)    & 0.38 & 0.84 & 1.32 & \multicolumn{1}{c|}{0.84} & \textbf{0.03} & \textbf{0.08} & \textbf{0.20} & \textbf{0.11} & \textbf{41.17}                   &   \textbf{11.36}                \\
Ours (UniAD \&. MiniCPM-V) & \textbf{0.32}     & \underline{0.67}     &  \textbf{1.08}    & \multicolumn{1}{c|}{\textbf{0.69}}     & 0.39     &  1.22    & 2.58     & 1.40     &   39.13                 &    10.00               \\
Ours (VAD \& Qwen-VL)      & 0.41     & 0.84     & 1.37     & \multicolumn{1}{c|}{0.87}     & \underline{0.06}    & \underline{0.15}     & \textbf{0.25}     & \underline{0.15 }    & \underline{40.76}                   &   \underline{10.93}                \\
Ours (UniAD \& Qwen-VL)    & \underline{ 0.33}    & \textbf{0.65}     & \underline{1.12}     & \multicolumn{1}{c|}{\underline{0.70}}     & 0.37     & 1.14     &  2.53    & 1.34     &   38.87                 &   10.00                \\ \bottomrule
\end{tabular}}
\vspace{-2mm}
\caption{\textbf{Open-loop and Closed-loop Planning Results on the Bench2Drive dataset. }
We reimplement models using the official code provided by the Bench2Drive paper. In the code, TCP plans for the next two seconds, while the other methods plan for the next three seconds. The best result is shown in \textbf{bold}, while the second-best is \underline{underlined}.}
\label{tab:bench2drive_comparison}
\vspace{-5mm}
\end{table*}

\begin{table*}[t]
\centering
\begin{tabular}{c|ccc|cccc|cccc}
\toprule
\multirow{2}{*}{Setting} &
  \multirow{2}{*}{SRI} &
  \multirow{2}{*}{TRI} &
  \multirow{2}{*}{HTD} &
  \multicolumn{4}{c|}{L2 (m) $\downarrow$} &
  \multicolumn{4}{c}{Collision Rate (\%) $\downarrow$} \\
  &    &    &    & 1s   & 2s   & 3s   & Avg. & 1s   & 2s   & 3s   & Avg. \\ 
\midrule
0 &    &    &    & 0.51 & 0.98 & 1.55 & 1.01 & 0.13 & 0.19 & 0.31 & 0.21 \\
1 & \checkmark &    &    & 0.44 & 0.86 & \textbf{1.36} & 0.89 & 0.08 & 0.16 & 0.26 & 0.17 \\
2 &    & \checkmark &    & 0.52 & 0.93 & 1.45 & 0.97 & 0.08 & 0.14 & 0.27 & 0.16 \\
3 &    &    & \checkmark & 0.46 & 0.89 & 1.46 & 0.93 & 0.05 & 0.16 & 0.21 & 0.14 \\
4 &\checkmark &\checkmark & &0.54 &0.92 &1.39 &0.95 &0.05  &\textbf{0.11} &0.29 &0.15 \\
5 &\checkmark & &\checkmark &0.42 &0.86 &1.39 &0.89 &0.08 &0.14 &0.23 &0.15 \\
6 & &\checkmark &\checkmark &0.44 &0.93 &1.50 &0.96 &0.08 &\textbf{0.11} &0.24 &0.14 \\
7 & \checkmark & \checkmark & \checkmark & \textbf{0.41} & \textbf{0.85} & 1.40 & \textbf{0.89} & \textbf{0.04} & 0.13 & \textbf{0.20} & \textbf{0.12} \\
\bottomrule
\end{tabular}
\vspace{-2mm}
\caption{Ablation study of each module on Bench2Drive dataset. ``SRI'' refers to Strategic Reasoning Injector module, ``TRI'' refers to Tactical Reasoning Integrator module and ``HTD'' means Hierarchical Trajectory Decoder module. The best result is in \textbf{bold}}.
\label{tab:ablation}
\vspace{-6mm}
\end{table*}

\subsection{Settings}
\textbf{Baseline. }
Our method supports different end-to-end planning networks and VLMs. In this paper, we use VAD \cite{jiang2023vad} and UniAD \cite{hu2023planning} as the baselines, and select MiniCPM-Llama3-2.5V \cite{yu2024rlaifv} and Qwen-VL \cite{bai2023qwen} as VLMs.

\noindent\textbf{Datasets. }
We evaluate open-loop planning on the nuScenes dataset \cite{caesar2020nuscenes}, which contains 1,000 20-second scenes annotated at 2Hz and serves as a key benchmark for E2E autonomous driving. For both open-loop and closed-loop evaluation, we use Bench2Drive \cite{jia2024bench2drive}, featuring 2M frames from 13,638 clips across 44 scenarios, 23 weather conditions, and 12 CARLA v2 towns. Its rigorous closed-loop protocol assesses E2E-AD models on 220 routes, ensuring a fair and comprehensive performance evaluation.

\noindent\textbf{Evaluation Metrics. }
For open-loop evaluation, we use L2 error and collision rate. L2 error measures the distance between planned and ground truth trajectories, while collision rate quantifies collisions with traffic participants. By default, we evaluate at 1, 2, and 3 seconds using VAD metrics \cite{jiang2023vad}. For closed-loop evaluation, following \cite{jia2024bench2drive}, we use Driving Score and Success Rate. Driving Score reflects route completion with infractions, while Success Rate is the percentage of routes completed without violations.

\noindent\textbf{Implementation Details. }
For baselines, we use the official codes and adhere to the hyperparameters specified in their official implementations. $\lambda_{0}$, $\lambda_{1}$, $\lambda_{2}$, $\lambda_{3}$ are set as 1.0, 0.5, 0.5 and 1.0. $\beta_c$ is 0.5 and $\beta_f$ is 1.0. Models are trained on 8 NVIDIA A40 GPUs or 4 NVIDIA 4090 GPUs using the PyTorch framework.

\subsection{Main Results}

\textbf{Open-loop evaluation. }
To better assess the effectiveness of our framework, we compare our method with several SOTA methods on nuScenes and Bench2Drive datasets. To better evaluate the effectiveness of our framework, we conducted experiments on two datasets, nuScenes and Bench2Drive, and compared it with several state-of-the-art methods. As shown in Tab.~\ref{tab:nuscenes_comparison} and ~\ref{tab:bench2drive_comparison}, our method achieves significant improvements over our baseline methods VAD and UniAD, especially in terms of L2 error and collision rate, which can reach more than 30\% improvement. It is worth noting that our performance is also better than other VLM-assessed methods such as VLP and VLM-AD that use the same baseline, achieving the lowest average L2 error (0.48m) and collision rate (0.15\%) on nuScenes, and 0.84m and 0.12\% on Bench2Drive. This shows that the introduction of human decision-making process makes the network more effective in learning driving ability.

\noindent\textbf{Closed-loop evaluation. }
Although the open-loop metrics provide partial performance results, we conduct closed-loop evaluation on Bench2Drive to assess real-world applicability. The results show that after integrating our framework, both the driving score and the number of completed paths have been significantly improved compared to the baseline, indicating that the introduction of human thinking processes has successfully improved driving mastery.

\subsection{Ablation Study}
We conduct an ablation study on the Bench2Drive validation set to evaluate our proposed modules. Refer to VAD, we adopt the two-stage training strategy here to accelerate experimentation. All ablation models share the same stage-1 checkpoint for fair comparison and all experiments use NVIDIA 4090 GPUs with VAD-base and MiniCPM-Llama3-2.5V baselines.

\noindent\textbf{Effectiveness of Strategic Reasoning Injector. }
To assess the effectiveness of the Strategic Reasoning Injector module, we conducted an ablation study by removing this component. As shown in Table~\ref{tab:ablation} (settings 0 and 1), its absence led to an approximately 12\% increase in average L2 error and a 19\% increase in average collision rate. These highlight that strategic decision-making guides ego-query modeling, allowing it to adaptively prioritize information relevant to the current scene decision-making process, improving overall planning performance.

\noindent\textbf{Effectiveness of Tactical Reasoning Integrator. }
As shown in Table~\ref{tab:ablation}, Settings 2 and 6 validate the effectiveness of this module. In Setting 2, we concatenate the embedded command features with the modeled ego-query features, and the final output is obtained through the decoder. Comparing Settings 0 and 2, we observe a 0.14m reduction in average L2 error and a 0.05\% decrease in average collision rate. These results suggest that tactical commands provide guidance that is closer to planning than strategic decisions, reducing the complexity of the learning space and enabling the network to make more informed decisions.

\noindent\textbf{Effectiveness of Hierarchical Trajectory Decoder. }
To highlight the importance of the Hierarchical Trajectory Decoder, we replace it with an MLP that directly predicts future trajectory based on ego-features. This modification leads to a 0.07m increase in L2 error and a 0.07\% rise in the collision rate (Setting 0 vs. 3), demonstrating the challenges of directly decoding fine-grained trajectories. The absence of hierarchical decoding—from easy to difficult and coarse to fine—hinders the model’s ability to refine trajectory predictions, ultimately suboptimal performance.

\subsection{More Analysis}
In this section, we further analyze our designed module on the Bench2Drive validation set. Settings are consistent with those in the ablation study. More \textbf{analytical experiments} and \textbf{visualization} for qualitative assessment of performance and interpretability, are provided in the \textcolor{red}{Appendix}.

\noindent\textbf{Dissusion on the use of similarity loss for encoded driving strategy features. } 
As shown in Tab.~\ref{tab:dissusion_sim_loss}, we removed the component of encoding ground truth trajectory and the similarity loss. We can observe that both L2 loss and collision rate increase, which can be attributed to the gap between the strategic strategy text features encoded by the VLM text encoder and the perception features used in the end-to-end network. The similarity loss helps the adapter bridge this gap, aligning the text features more closely with those required for trajectory prediction. This, in turn, better guides the learning of ego-query and improves performance.

\noindent\textbf{Dissusion on different numbers of layers in the hierarchical trajectory decoder.}
As shown in Tab.~\ref{tab:dissusion_layers_in_decoder}, we experiment with different numbers of layers in decoder. One layer represents the direct output of the fine-grained trajectory, while two layers follow the described methods. In the three-layer setting, we adjust the decoder to three layers and use Bezier curve fitting on the coarse trajectory to generate a coarser ground truth for joint supervision. Experiments show that increasing from one to two layers significantly improves performance, but too many layers introduce unnecessary complexity, leading to overfitting and loss of fine-grained details critical for accurate prediction.

\begin{table}[]
\centering
\resizebox{\linewidth}{!}{
\begin{tabular}{l|cccc|cccc}
\toprule
\multirow{2}{*}{Model}          & \multicolumn{4}{c|}{L2(m) $\downarrow$} & \multicolumn{4}{c}{Collision Rate (\%) $\downarrow$} \\
                                & 1s    & 2s   & 3s   & Avg. & 1s     & 2s     & 3s     & Avg.   \\ \midrule
w/o loss &0.45&0.88&1.43&0.92 &0.09&0.16&0.27& 0.18\\
w. loss & \textbf{0.44} & \textbf{0.86} & \textbf{1.36} & \textbf{0.89} & \textbf{0.08} & 0.16 & \textbf{0.26} & \textbf{0.17} \\
\bottomrule
\end{tabular}}
\vspace{-2mm}
\caption{Ablation study to evaluate the use of similarity loss for encoded driving strategy features.}
\label{tab:dissusion_sim_loss}
\vspace{-3mm}
\end{table}

\begin{table}[]
\centering
\resizebox{\linewidth}{!}{
\begin{tabular}{c|cccc|cccc}
\toprule
\multirow{2}{*}{\#layers} & \multicolumn{4}{c|}{L2(m) $\downarrow$} & \multicolumn{4}{c}{Collision Rate (\%) $\downarrow$} \\
                          & 1s    & 2s   & 3s   & Avg. & 1s       & 2s       & 3s      & Avg.    \\ \midrule
1   & 0.51 &0.95 &1.51 &0.99 & 0.07 &\textbf{0.15} &0.30 &0.17  \\
2   & \textbf{0.46} & \textbf{0.89} &\textbf{1.46} & \textbf{0.93} & \textbf{0.05} & 0.16 & \textbf{0.21} & \textbf{0.14} \\
3   & 0.48 & 0.93 &1.52 & 0.97 & \textbf{0.05} & 0.17 & 0.24 & 0.15    \\ \bottomrule
\end{tabular}}
\vspace{-2mm}
\caption{Ablation study for different numbers of layers in the hierarchical trajectory decoder.}
\label{tab:dissusion_layers_in_decoder}
\vspace{-6mm}
\end{table}

\section{Conclusions}
\label{sec:conclusions}
In conclusion, our paper presents ReAL-AD, a reasoning-augmented learning framework that enhances end-to-end autonomous driving by leveraging Vision-Language Models (VLMs) for structured reasoning across strategy, decision, and operation levels. By modeling human-like hierarchical decision-making, ReAL-AD integrates strategic decisions with tactical commands and trajectory refinement. Extensive experiments on NuScenes and Bench2Drive demonstrate its state-of-the-art performance in trajectory planning accuracy and driving safety.

\section{Acknowledgement}
This work was supported by NSFC (No.62206173), Shanghai Frontiers Science Center of Human-centered Artificial Intelligence (ShangHAI), and MoE Key Laboratory of Intelligent Perception and Human-Machine Collaboration (KLIP-HuMaCo).

{
    \small
    \bibliographystyle{ieeenat_fullname}
    \bibliography{main}
}

\clearpage
\appendix
\noindent \textbf{\Large Appendix} \\

In the appendix, we first present additional model details, followed by analytical experiments to further demonstrate our model's effectiveness, and finally we provide visualization results.

\section{More details about method}
In this section, we provide more details of our proposed ReAL-AD framework, covering the Strategic Reasoning Injector (SRI) module, the Tactical Reasoning Integrator (TRI) module, the Hierarchical Trajectory Decoder (HTD) module, and model adaptations for closed-loop evaluation.

\subsection{More details of SRI module}
\textbf{Driving Strategy Prompt. }The prompt used to generate driving strategy text is as follow:

\begin{tcolorbox}[
    title={\normalfont\textbf{Strategy Prompt}},
    colback=white,
    colframe=black!50,
    fonttitle=\bfseries,
    boxsep=3pt,
    left=6pt,
    right=6pt,  
    arc=0mm,     
    boxrule=0.5pt,  
    before upper={\parindent0pt},
    width=\columnwidth,
]
There is an image from front camera of the car. Suppose you are driving in this scenario, based on the scene condition and critical objects or traffic signs in the scene, make legal, safe and comfortable driving decisions (may include which traffic participants to pay attention to, which traffic rules to pay attention to, etc.). Answer should be in one paragraph and the format is as follows: 

\noindent
\ttfamily
'When driving in the current scenario, the driving decision-making thinking process is:...'.
\end{tcolorbox}

\noindent\textbf{Details about ground truth trajectory encoder. }
In the SRI module, we encode the ground truth trajectory into planning features to ensure consistency between strategy semantics and planning dynamics. The pseudo-code for this encoder is presented in Algorithm \ref{code:gt_encoder}.

\begin{algorithm}
\caption{Ground-truth trajectory encoder}
\begin{algorithmic}[1]

\State \textbf{Input:} Ground truth trajectory $gt_{traj}$ of shape $(B, N, T, D)$
\State \textbf{Hyperparameters:} Number of subgraph layers $L$, Hidden dimension $H$
\State \textbf{Output:} Extracted trajectory features $F_{\text{gt}}$

\State Initialize feature representation $F \gets gt_{traj}$

\For{$i = 1$ to $L$} 
    \State $F \gets \text{MLP}_i(F)$ \Comment{Apply MLP to each point feature}
    \State $F_{\text{max}} \gets \max(F, \text{axis}=-2)$ \Comment{Max pooling along sequence length}
    \State $F_{\text{max\_expanded}} \gets \text{Expand}(F_{\text{max}}, \text{axis}=-2, \text{size}=T)$
    \State $F \gets \text{Concat}(F, F_{\text{max\_expanded}}, \text{axis}=-1)$ \Comment{Feature concatenation}
\EndFor

\State $F_{\text{gt}} \gets \max(F, \text{axis}=-2)$ \Comment{Final max pooling for global trajectory features}

\State \Return $F_{\text{gt}}$

\end{algorithmic}
\label{code:gt_encoder}
\end{algorithm}

Beyond ensuring consistency, aligning the encoded ground truth trajectory with the loss of the driving strategy features provides additional benefits. In rare instances where the driving strategy inferred by the VLM conflicts with the generated driving command, the GT trajectory serves as a common supervisory signal for both. Specifically, the loss between the driving strategy features and the GT trajectory helps refine the strategy representation, while the generated driving command, once injected into the decoder, is also supervised by the correct trajectory at both coarse and fine-grained levels. Since both the strategy features and the command share this trajectory-based supervision, the model benefits from a unified correction mechanism, effectively mitigating the impact of inconsistencies in the VLM’s output.

\noindent\textbf{Verification of VLM output strategy text.}
Due to differences in command-following capabilities among VLMs, they may not always adhere to the specified output format, even when explicitly instructed. To ensure the reliability of VLM-generated strategy text, we rigorously verify that the response begins with ``\textit{When driving in the current scenario...}" as specified in the prompt. If it does not, we regenerate the response.

\subsection{More details of TRI module}
\noindent\textbf{Driving Command Prompt.}
To generate tactical command, we design the prompt as Fig.~\ref{promptbox:command_prompt} shows.

\begin{figure*}[t]
\centering
\begin{tcolorbox}[
    title={\normalfont\textbf{Command Prompt}},
    colback=white,
    colframe=black!50,
    fonttitle=\bfseries,
    boxsep=3pt,
    left=6pt,
    right=6pt,
    arc=0mm,     
    boxrule=0.5pt,  
    width=\textwidth 
]
There is an image from the front camera of the car. Suppose you are driving in this scenario, please analyze the following input image and determine the ego-vehicle's planning commands based on the current road conditions. Strictly adhere to the predefined command types listed below and return the results in the required format. The output must only include the specified options and should not contain any additional content.

\medskip
\textbf{Command Types}

\noindent
\textit{Primary Direction Control}
\begin{itemize}[nosep,leftmargin=12pt]
    \item LEFT\_TURN: Make a left turn at the upcoming intersection or turn
    \item RIGHT\_TURN: Make a right turn at the upcoming intersection or turn
    \item CONTINUE\_STRAIGHT: Proceed straight without changing direction. Maintain the current lane unless a lane change is necessary for traffic flow
\end{itemize}

\noindent
\textit{Lane Positioning}
\begin{itemize}[nosep,leftmargin=12pt]
    \item KEEP\_LANE: Maintain the current lane without changing
    \item CHANGE\_LANE\_LEFT: Move to the adjacent lane on the left to facilitate a left turn or overtake slower traffic
    \item CHANGE\_LANE\_RIGHT: Move to the adjacent lane on the right to facilitate a right turn or overtake slower traffic
\end{itemize}

\noindent
\textit{Speed Regulation}
\begin{itemize}[nosep,leftmargin=12pt]
    \item ACCELERATE: Increase the vehicle's speed
    \item DECELERATE: Decrease the vehicle's speed
    \item MAINTAIN\_SPEED: Keep the current speed constant
\end{itemize}

\noindent
\textit{Emergency Control}
\begin{itemize}[nosep,leftmargin=12pt]
    \item EMERGENCY\_BRAKE: Apply brakes immediately to avoid a collision or hazard
    \item PARK: Bring the vehicle to a complete stop and park
    \item NO\_ACTION: No emergency control action is required in the current scenario
\end{itemize}

\vspace{0.5em}
\noindent
\textbf{Output Format}

\begin{lstlisting}[basicstyle=\small\ttfamily,frame=leftline,framesep=5pt]
Direction Control: <Direction Control>
Lane Management: <Lane Management>
Speed Control: <Speed Control>
Emergency Control: <Emergency Control>
\end{lstlisting}
\end{tcolorbox}
\caption{Command prompt structure for tactical decision-making.}
\label{promptbox:command_prompt}

\end{figure*}

\noindent\textbf{Verification of VLM output driving command.}
Similar to strategy text, we use regular expressions to extract structures such as ``\textit{Direction Control:}" and three other similar categories from the text, identifying the word following each as the command reply. Additionally, each category has a predefined set of valid responses. If the VLM output does not match any of the expected answers within its respective category or if the structure is missing, the output is regenerated.

\subsection{More details of HTD module.}
\noindent\textbf{Coarse trajectory ground truth generation. }
We convert the fine trajectory ground truth into a coarse trajectory through Bézier curve interpolation~\cite{pastva1998bezier} and strategic control point sampling. This process consists of four key stages.

First, we transform the offset coordinates $\{\Delta\mathbf{P}_i\}_{i=1}^6 \in \mathbb{R}^{6\times2}$ into absolute coordinates via cumulative summation:
\begin{equation}
    \mathbf{P}^{abs}_i = 
    \begin{cases} 
        \Delta\mathbf{P}_1, & i=1 \\
        \mathbf{P}^{abs}_{i-1} + \Delta\mathbf{P}_i, & 2 \leq i \leq 6
    \end{cases}
\end{equation}
This step ensures spatial continuity, which is crucial for the subsequent interpolation.

Next, we extract four key control points from $\{\mathbf{P}^{abs}_i\}$ to effectively capture kinematic patterns while maintaining smoothness. Specifically, we select $\mathbf{P}^{abs}_1$ as the starting point, $\mathbf{P}^{abs}_3$ and $\mathbf{P}^{abs}_4$ as intermediate points representing motion inflection regions, and $\mathbf{P}^{abs}_6$ as the endpoint.

To obtain a smoothed trajectory, we then apply cubic Bézier interpolation using these control points. The trajectory $\{\mathbf{\widetilde{P}}^{abs}_i\} \in \mathbb{R}^{6\times2}$ is computed as:
\begin{equation}
    \mathbf{B}(t) = \sum_{k=0}^{3} \mathbf{C}_{k+1} \binom{3}{k} t^k (1-t)^{3-k}, \quad t \in [0,1]
\end{equation}
where we uniformly sample at $t = \{0, 0.2, 0.4, 0.6, 0.8, 1\}$ to maintain temporal consistency with the original trajectory.

Finally, we convert the interpolated absolute coordinates back to the offset form:
\begin{equation}
    \Delta\mathbf{\widetilde{P}}_i = 
    \begin{cases} 
        \mathbf{\widetilde{P}}^{abs}_1, & i=1 \\
        \mathbf{\widetilde{P}}^{abs}_i - \mathbf{\widetilde{P}}^{abs}_{i-1}, & 2 \leq i \leq 6
    \end{cases}
\end{equation}
The resulting coarse trajectory $\{\Delta\mathbf{\widetilde{P}}_i\}$ preserves the global motion trends and serves as supervision for the network's predicted trajectory.

\noindent\textbf{Distribution Encoder. }
In this module, we leverage Distribution Encoder to map decoder input into latent trajectory space. The operation is as follow:

\begin{algorithm}
\caption{Distribution Encoder}
\begin{algorithmic}[1]

\State \textbf{Input:} Input tensor $s_t$, consisting of concatenated ego-vehicle features, corresponding command features, and optionally coarse latent code, with shape $(B, T, D)$.
\State \textbf{Hyperparameters:} Latent dimension $L$, Minimum log variance $\text{min\_log\_sigma}$, Maximum log variance $\text{max\_log\_sigma}$
\State \textbf{Output:} Mean $\mu$ and log standard deviation $\log \sigma$

\State Initialize compression dimension $C \gets D / 2$
\State Initialize encoder $E$ with three convolutional layers:
\State \hspace{1em} $\text{Conv1}: (D \to 2D, 1 \times 1)$, ReLU
\State \hspace{1em} $\text{Conv2}: (2D \to 2D, 1 \times 1)$, ReLU
\State \hspace{1em} $\text{Conv3}: (2D \to C, 1 \times 1)$

\State Initialize final convolutional layer $\text{LastConv}$:
\State \hspace{1em} Adaptive average pooling $(T \to 1)$
\State \hspace{1em} $\text{Conv}: (C \to 2L, 1 \times 1)$

\State \textbf{Encoding Process:}
\State $s_t \gets \text{Permute}(s_t, (0,2,1))$ 
\State $F \gets E(s_t)$ 
\State $mu\_log\_sigma \gets \text{LastConv}(F)$
\State $mu\_log\_sigma \gets \text{Permute}(mu\_log\_sigma, (0,2,1))$ 

\State \textbf{Extract Distribution Parameters:}
\State $\mu \gets mu\_log\_sigma[:, :, :L]$
\State $\log\sigma \gets mu\_log\_sigma[:, :, L:]$
\State $\log\sigma \gets \text{Clamp}(\log\sigma, \text{min\_log\_sigma}, \text{max\_log\_sigma})$

\State \Return $\mu, \log\sigma$

\end{algorithmic}
\end{algorithm}

\subsection{Model adaptation for closed-loop evaluation}
For closed-loop evaluation, we adapted the model to improve processing speed, particularly for the slow VLM network. To reduce VLM usage frequency and increase FPS, we implemented a memory buffer mechanism that stores environmental features along with the corresponding VLM policy text and driving instruction features from the open-loop evaluation dataset. During closed-loop evaluation, when encountering a new scene, the system first extracts environmental features and computes their similarity with those in the buffer. It then identifies the most similar stored frame based on environmental feature similarity and reuses its VLM policy text and driving instruction features for the current frame, ensuring efficient decision-making.

\section{More Analysis}
In this section, we will provide more analytical experiments on our proposed modules. All experimental settings are consistent with those in the ablation study of main paper.

\subsection{Discussion on design of the adapter in SRI module. }
In the SRI module, the adapter aligns encoded strategy textual features with perceptual features. While our main implementation uses an MLP, we also experimented with a self-attention mechanism. However, Tab.~\ref{tab:different_adapter_in_SRI} shows no significant performance difference. This suggests that the MLP is already sufficient for feature adaptation, as the alignment task does not heavily rely on long-range dependencies. Additionally, the MLP is more computationally efficient, making it a practical choice.

\begin{table}[]
\centering
\resizebox{\linewidth}{!}{
\begin{tabular}{l|cccc|cccc}
\toprule
\multirow{2}{*}{Sim Loss} & \multicolumn{4}{c|}{L2(m) $\downarrow$} & \multicolumn{4}{c}{Collision Rate (\%) $\downarrow$} \\
         & 1s   & 2s   & 3s   & Avg. & 1s   & 2s   & 3s   & Avg. \\ \midrule
SelfAttn & 0.43 & 0.87 & 1.40 & 0.90 & 0.08 & 0.17 & 0.24 & 0.16 \\
MLP      & 0.44 & 0.86 & 1.36 & 0.89 & 0.08 & 0.16 & 0.26 & 0.17 \\ \bottomrule
\end{tabular}}
\caption{Ablation study on different adapter designs in the SRI module.}
\label{tab:different_adapter_in_SRI}
\end{table}

\subsection{Discussion on similarity loss selection in SRI module. }
To investigate the impact of different similarity loss functions in the SRI module, we conduct an ablation study comparing Cosine Similarity Loss (our default choice) with Mean Squared Error (MSE) Loss and Kullback-Leibler (KL) Divergence Loss. The results, as shown in Tab.~\ref{tab:Discussion_type_of_sim_loss}, indicate that MSE Loss performs the worst, likely due to its sensitivity to magnitude differences, which can lead to instability when aligning high-dimensional feature spaces. Meanwhile, KL Divergence Loss and Cosine Similarity Loss achieve comparable performance, suggesting that capturing relative distribution differences (as in KL Loss) or directional alignment (as in Cosine Loss) is more effective than direct magnitude-based regression.

These findings reinforce that the specific choice of loss function is not the primary focus of our design. Instead, the core contribution lies in the conceptual framework of aligning these two features effectively, ensuring consistency between strategy semantics and planning dynamics.

\begin{table}[]
\centering
\resizebox{\linewidth}{!}{
\begin{tabular}{l|cccc|cccc}
\toprule
\multirow{2}{*}{Loss Function} & \multicolumn{4}{c|}{L2(m) $\downarrow$} & \multicolumn{4}{c}{Collision Rate (\%) $\downarrow$} \\
            & 1s   & 2s   & 3s   & Avg. & 1s   & 2s   & 3s   & Avg. \\ \midrule
None        & 0.45 & 0.88 & 1.43 & 0.92 & 0.09 & 0.16 & 0.27 & 0.18 \\
MSE Loss    & 0.47 & 0.89 & 1.45 & 0.93 & 0.11 & 0.16 & 0.32 & 0.20 \\
KL Loss     & 0.43 & 0.88 & 1.35 & 0.89 & 0.07 & 0.19 & 0.28 & 0.18 \\
CosSim Loss& 0.44 & 0.86 & 1.36 & 0.89 & 0.08 & 0.16 & 0.26 & 0.17 \\ \bottomrule
\end{tabular}}
\caption{Ablation study to different similarity loss functions in the SRI module.}
\label{tab:Discussion_type_of_sim_loss}
\end{table}

\subsection{Discussion on different forms of tactical driving commands.}
As shown in Tab.\ref{tab:Discussion_form_of_command}, we modify the prompt input to the VLM to generate tactical commands in a probabilistic form (see Fig.\ref{promptbox:command_prompt_prob}) and encode them using a MLP, while keeping all other settings unchanged. However, this adjustment results in a performance decline. The probabilistic representation introduces uncertainty, making it more challenging for the model to produce clear, decisive predictions. In contrast, the one-hot format offers more direct supervision, leading to improved performance.

\begin{table}[]
\centering
\resizebox{\linewidth}{!}{
\begin{tabular}{l|cccc|cccc}
\toprule
\multirow{2}{*}{Model}          & \multicolumn{4}{c|}{L2(m) $\downarrow$} & \multicolumn{4}{c}{Collision Rate (\%) $\downarrow$} \\
                                & 1s    & 2s   & 3s   & Avg. & 1s     & 2s     & 3s     & Avg.   \\ \midrule
one-hot form & 0.52 & 0.93 & 1.45 & 0.97 & \textbf{0.08} & \textbf{0.14} & \textbf{0.27} & \textbf{0.16} \\
probabilistic form &\textbf{0.42} &\textbf{0.83} &\textbf{1.32} &\textbf{0.86} &0.11 &0.20 &0.28 &0.20 \\
\bottomrule
\end{tabular}}
\caption{Ablation study for different forms of tactical driving commands.}
\label{tab:Discussion_form_of_command}
\end{table}

\subsection{Discussion on camera input of VLMs.}
Our current design focuses on front-view reasoning to balance computational efficiency and performance, as strategic and tactical decisions (e.g., lane changes) primarily depend on the front environment. We acknowledge, however, that multi-view input can be beneficial in special cases. To explore this, we use VAD and Qwen-VL with multi-view input but get limited improvement, suggesting that most decisions can be accurately made using the front view alone. But multi-view setting is truly deserved to further explore.
\begin{table}[h]
\centering
\resizebox{\linewidth}{!}{
\begin{tabular}{lcc}
\toprule
Model                  & Avg. L2 (m) & Avg. Col. Rate (\%) \\ \midrule
Ours                   & 0.53        & 0.17                \\
Ours with multi-view input & 0.51        & 0.16                \\ \bottomrule
\end{tabular}}
\caption{Ablation study for different number of image input of VLM.}
\end{table}

\begin{figure*}[t]
\centering
\begin{tcolorbox}[
    title={\normalfont\textbf{Probability-based Command Prompt}},
    colback=white,
    colframe=black!50,
    fonttitle=\bfseries,
    boxsep=3pt,
    left=6pt,
    right=6pt,
    arc=0mm,     
    boxrule=0.5pt,  
    width=\textwidth 
]
There is an image from the front camera of the car. Suppose you are driving in this scenario, please analyze the following input image and determine the ego-vehicle's planning commands based on the current road conditions. **Strictly adhere to the predefined command types listed below and return the results in the required format. The output must **only** include the specified options and should **not** contain any additional content.**

\medskip
\noindent
\textbf{Important:}
\begin{enumerate}[nosep,leftmargin=12pt]
    \item You must provide a probability distribution for each command category, ensuring that the probabilities for each category sum up to 1.
    \item Use format strictly as shown below without any extra text or explanations.
    \item Refer to the provided examples to guide your output format and content.
\end{enumerate}

\medskip
\noindent
\textbf{Command Types}

\noindent
\textit{Emergency Control}
\begin{itemize}[nosep,leftmargin=12pt]
    \item \textbf{EMERGENCY\_BRAKE}: Immediately apply emergency brakes to avoid a collision or respond to a sudden hazard.
    \item \textbf{PARK}: Bring the vehicle to a complete stop and park, suitable for situations requiring immediate cessation.
    \item \textbf{NO\_ACTION}: No emergency control action is required in the current scenario, and the vehicle continues normal operation.
\end{itemize}

\noindent
\textit{Primary Direction Control}
\begin{itemize}[nosep,leftmargin=12pt]
    \item \textbf{LEFT\_TURN}: Make a left turn at the upcoming intersection or turn into another street.
    \item \textbf{RIGHT\_TURN}: Make a right turn at the upcoming intersection or turn into a parking area or another street.
    \item \textbf{CONTINUE\_STRAIGHT}: Proceed straight along the current route without changing direction.
\end{itemize}

\noindent
\textit{Lane Management}
\begin{itemize}[nosep,leftmargin=12pt]
    \item \textbf{KEEP\_LANE}: Maintain the current lane without changing, suitable for smooth traffic flow within the lane.
    \item \textbf{CHANGE\_LANE\_LEFT}: Move to the adjacent lane on the left, typically for overtaking or preparing for a left turn.
    \item \textbf{CHANGE\_LANE\_RIGHT}: Move to the adjacent lane on the right, typically for overtaking or preparing for a right turn.
\end{itemize}

\noindent
\textit{Speed Control}
\begin{itemize}[nosep,leftmargin=12pt]
    \item \textbf{ACCELERATE}: Increase the vehicle's speed, suitable for smooth traffic flow ahead or when overtaking.
    \item \textbf{DECELERATE}: Decrease the vehicle's speed, suitable for congested traffic, reduced speed limits, or obstacles ahead.
    \item \textbf{MAINTAIN\_SPEED}: Keep the current speed constant, suitable for stable traffic conditions or when no speed adjustment is needed.
\end{itemize}

\medskip
\noindent
\textbf{Output Format}

\begin{lstlisting}[basicstyle=\small\ttfamily,frame=leftline,framesep=5pt]
{
  "Emergency Control": {
    "EMERGENCY_BRAKE": <probability>,
    "PARK": <probability>,
    "NO_ACTION": <probability>
  },
  "Primary Direction Control": {
    "LEFT_TURN": <probability>,
    "RIGHT_TURN": <probability>,
    "CONTINUE_STRAIGHT": <probability>
  },
  "Lane Management": {
    "KEEP_LANE": <probability>,
    "CHANGE_LANE_LEFT": <probability>,
    "CHANGE_LANE_RIGHT": <probability>
  },
  "Speed Control": {
    "ACCELERATE": <probability>,
    "DECELERATE": <probability>,
    "MAINTAIN_SPEED": <probability>,
    "STOP": <probability>
  }
}
\end{lstlisting}

\end{tcolorbox}
\caption{Command prompt structure for tactical decision-making with probability-based outputs.}
\label{promptbox:command_prompt_prob}
\end{figure*}

\section{Visualization}
In this section, we present comprehensive visualization results of our approach.

\subsection{Qualitative results. }
\begin{figure*}[h]
  \includegraphics[width=\linewidth]{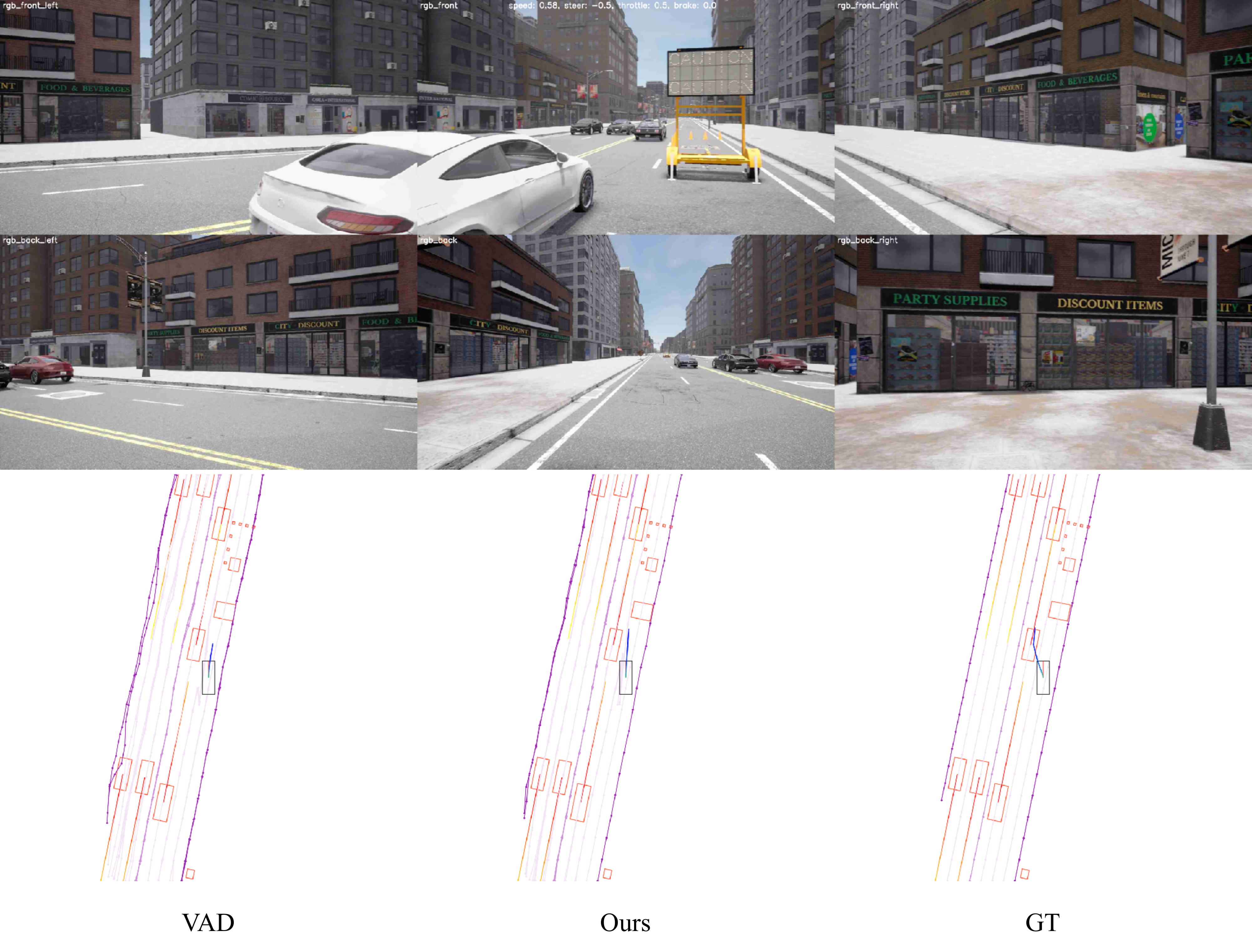}
  \caption{Visualization of our method, depicting a scenario where an error is detected in the current lane while driving, requiring a lane change to proceed.}
\label{fig:vis_construction_obstacle}
\end{figure*}

\begin{figure*}[h]
  \includegraphics[width=\linewidth]{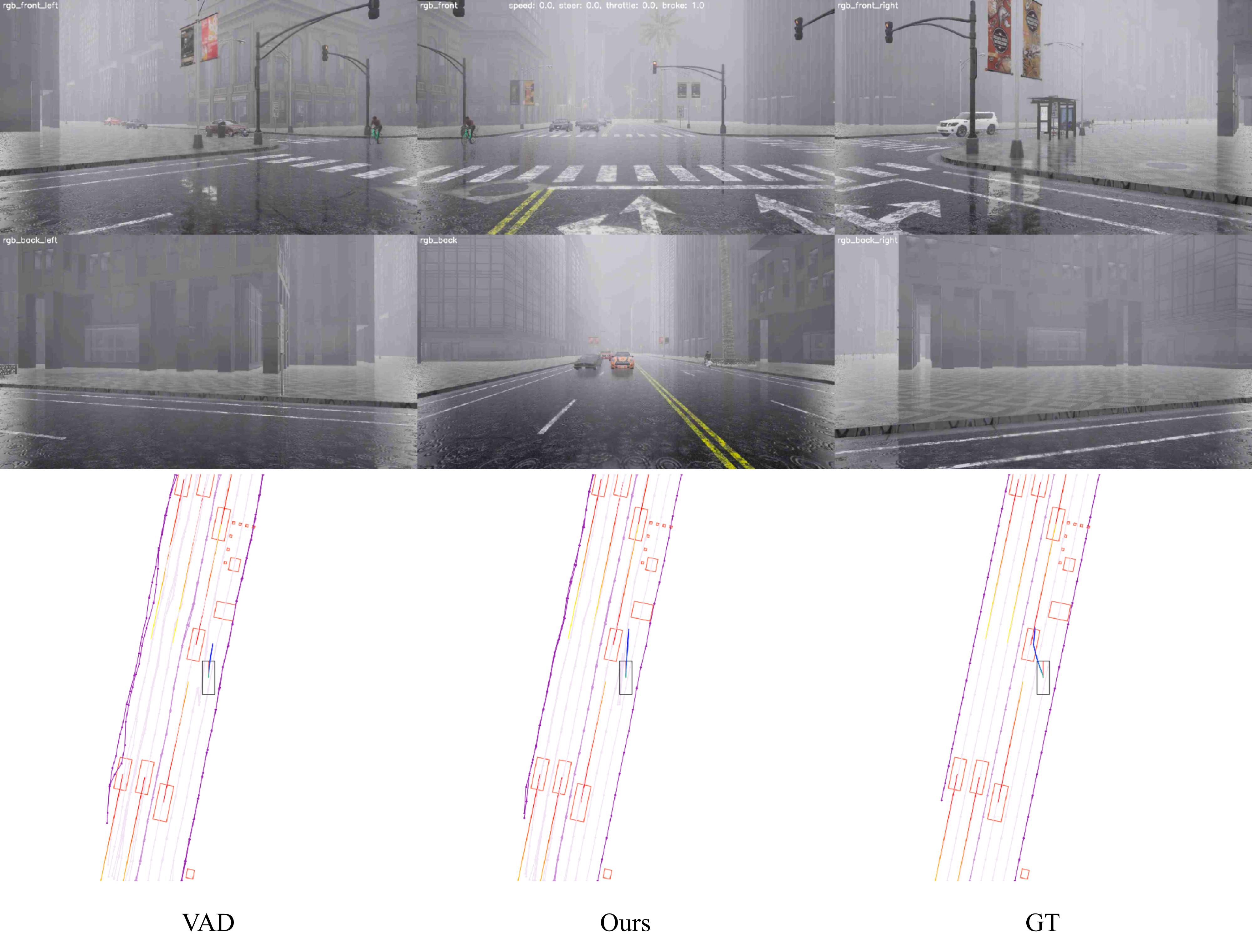}
  \caption{Visualization of our methods, depicting a scenario where a pedestrian suddenly appears in front of you and you need to make emergency avoidance while driving. }
\label{fig:vis_dynamic_object_crossing}
\end{figure*}

\begin{figure*}[h]
  \includegraphics[width=\linewidth]{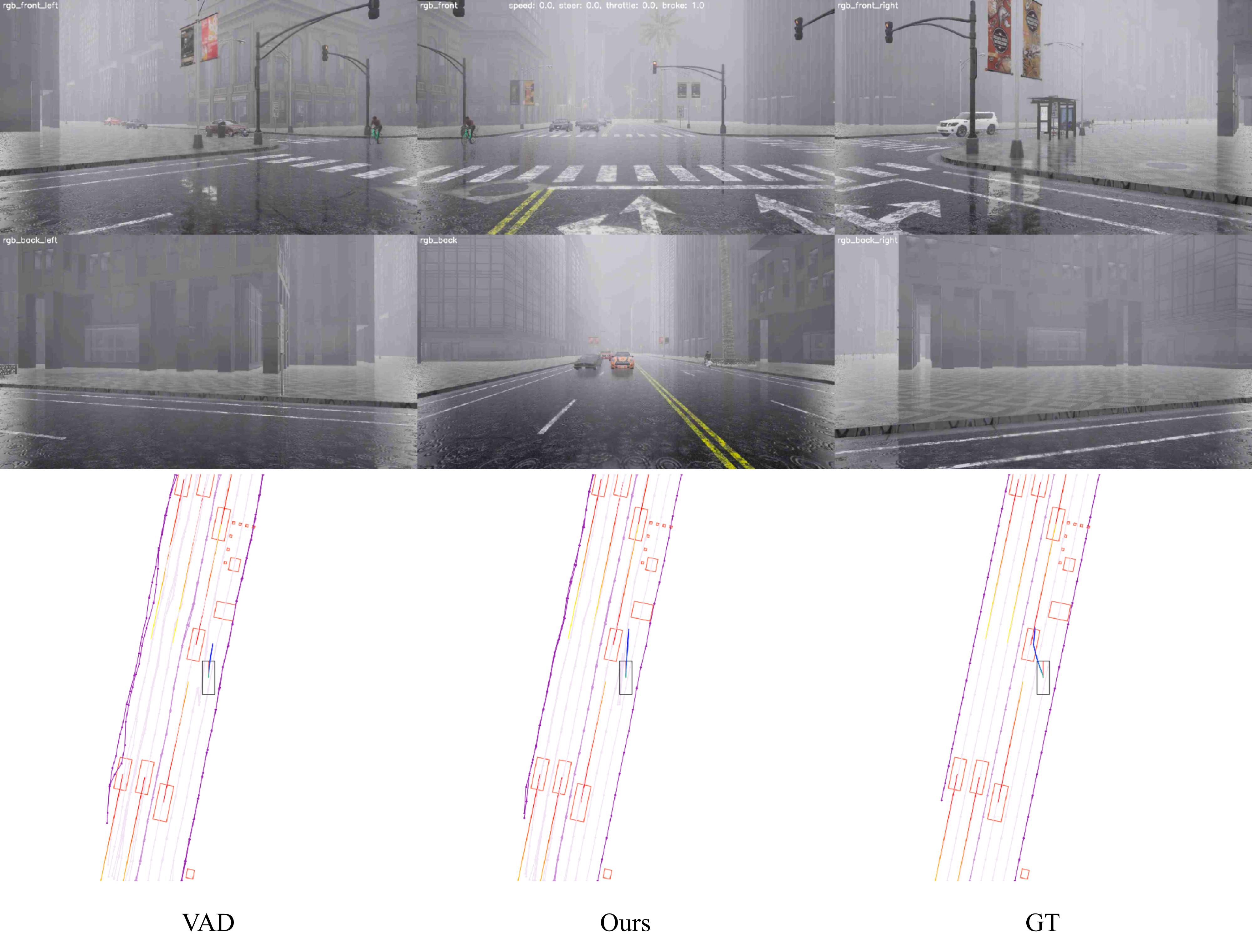}
  \caption{Visualization of our methods, depicting a turn without a signal.}
\label{fig:vis_non_signalized_turn_left}
\end{figure*}

\begin{figure*}[h]
  \includegraphics[width=\linewidth]{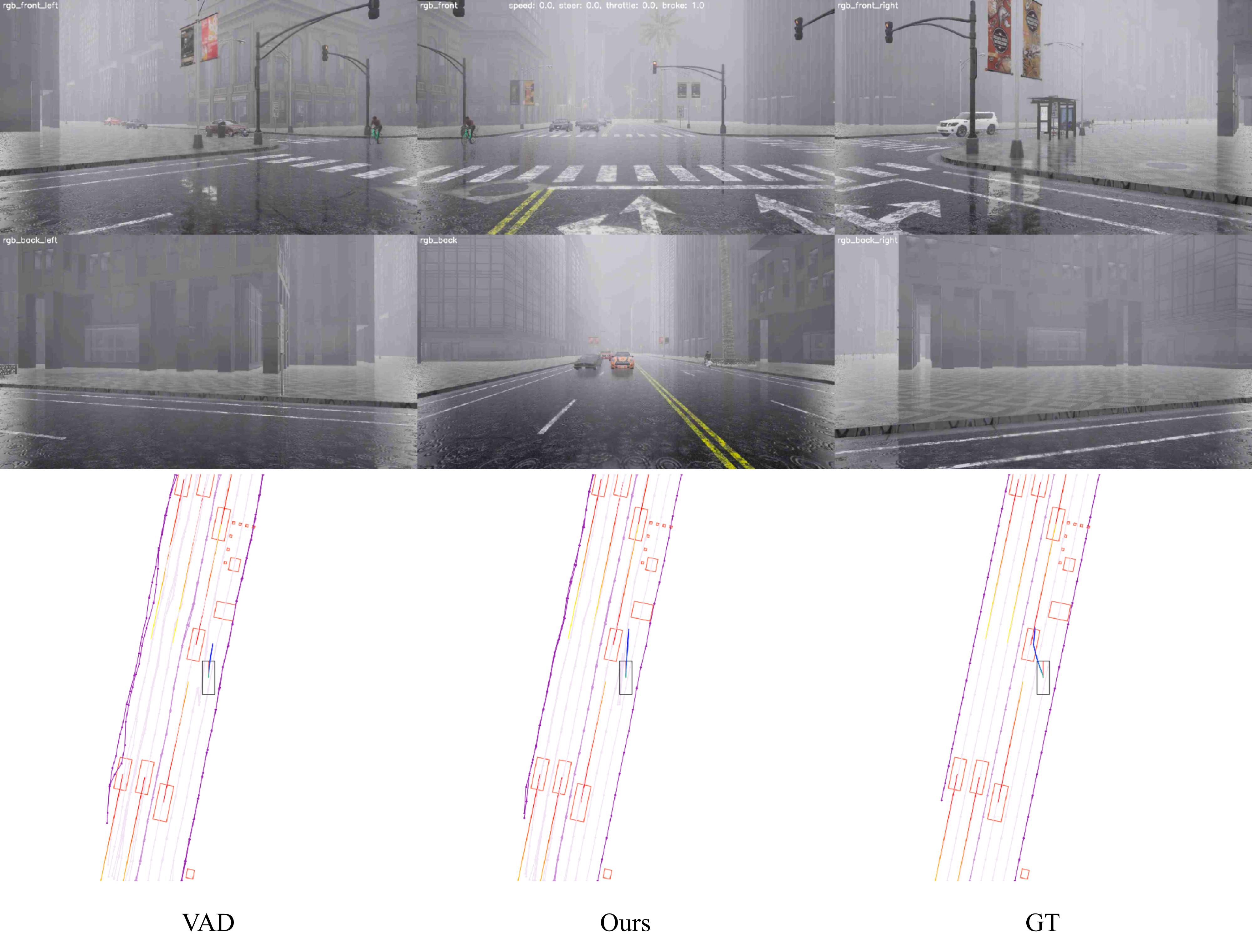}
  \caption{Visualization of our methods,  depicting a turn with a signal.}
\label{fig:vis_signalized_turn_right}
\end{figure*}

We use VAD~\cite{jiang2023vad} as the baseline and select MiniCPM-Llama3-2.5V~\cite{yu2024rlaifv} as the VLM, then perform a qualitative analysis of our proposed framework, as illustrated in Fig.~\ref{fig:vis_construction_obstacle},~\ref{fig:vis_dynamic_object_crossing},~\ref{fig:vis_non_signalized_turn_left} and ~\ref{fig:vis_signalized_turn_right}. In Fig.~\ref{fig:vis_construction_obstacle}, the baseline method failed to recognize the intention to change lanes, whereas our method accurately captured it and executed a safe lane change. In Fig.~\ref{fig:vis_dynamic_object_crossing}, when a pedestrian suddenly appeared, the baseline method responded too slowly and failed to perform an emergency avoidance maneuver, while our method successfully did so. In Fig.~\ref{fig:vis_non_signalized_turn_left} and \ref{fig:vis_signalized_turn_right}, the baseline method did not recognize the turning intention and missed the necessary turns, whereas our method correctly executed them. These results demonstrate that incorporating human reasoning into the end-to-end autonomous driving system enhances both the accuracy and safety of planning.

\subsection{Demonstration of enhanced interpretability. }
\begin{figure*}[h]
  \includegraphics[width=\linewidth]{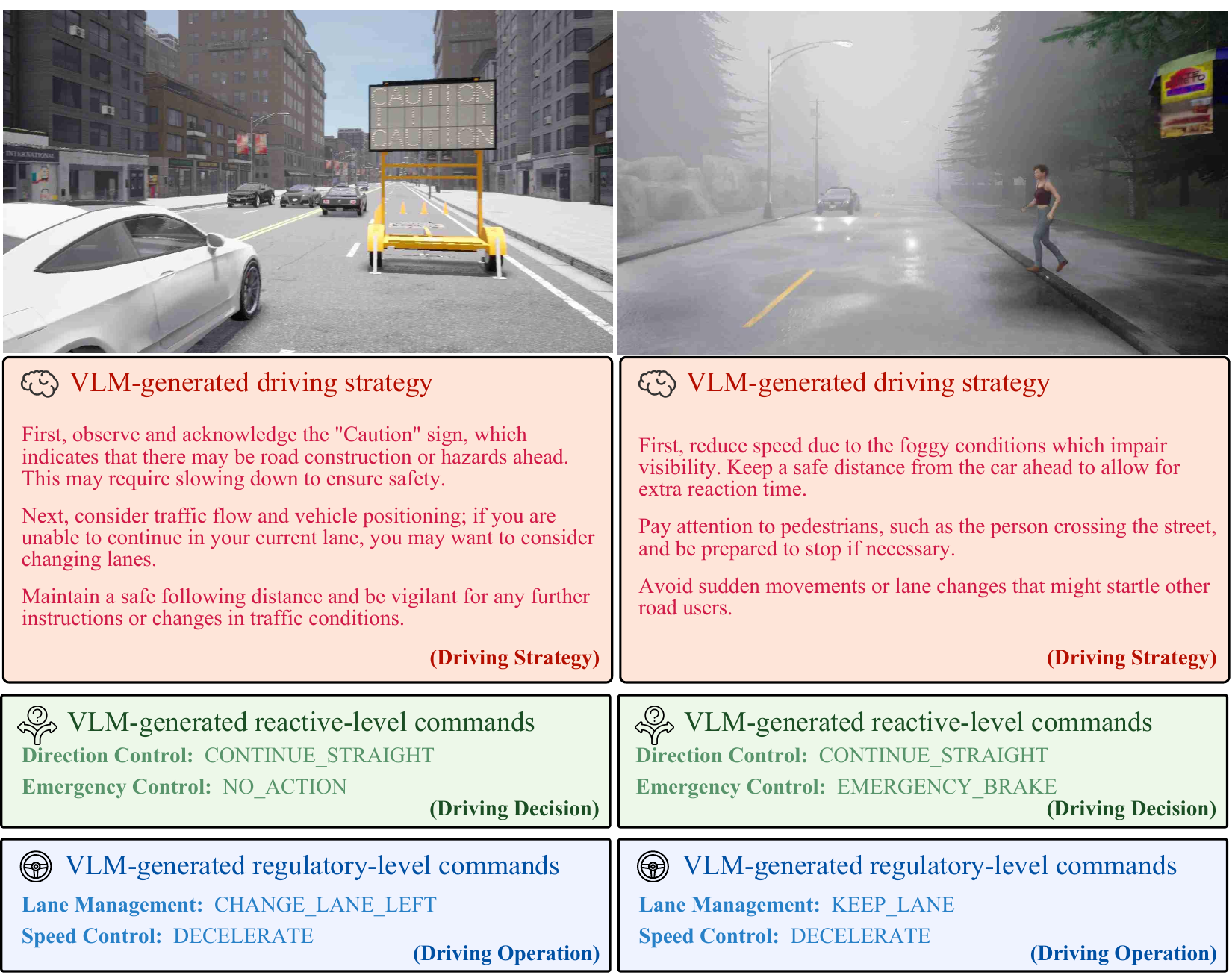}
  \caption{Visualization of VLM-generated driving strategies and tacical commands, demonstrating their alignment with final planning.}
\label{fig:show_interpretability}
\end{figure*}

Our proposed module enhances the interpretability of end-to-end trajectory prediction by incorporating human-like reasoning. Taking the two challenging scenarios in Fig.~\ref{fig:show_interpretability} as examples, we observe that the generated driving strategies align with the final trajectory predictions, reinforcing interpretability. In the left scenario, the model explicitly suggests a lane change to continue driving, which matches the command-level output. In the right scenario, the strategy advises slowing down and monitoring pedestrians, while the command recommends emergency braking—both contributing to safe trajectory planning. This alignment between intermediate reasoning and final predictions makes the decision-making process more transparent and interpretable.

\end{document}